%% file: main.tex
\documentclass[letterpaper]{article} 
\usepackage{aaai25}  
\usepackage{times}  
\usepackage{helvet}  
\usepackage{courier}  
\usepackage[hyphens]{url}  
\usepackage{graphicx} 
\usepackage{natbib}   
\usepackage{caption} 
\usepackage{algorithm}
\usepackage[noend]{algpseudocode}
\usepackage{cite}
\usepackage{amsmath}
\usepackage{amssymb}
\usepackage{multirow}  
\usepackage{booktabs}  
\usepackage{tabularx}  
\usepackage{subcaption}
\usepackage{siunitx}
\usepackage[table,dvipsnames]{xcolor}
\definecolor{softgreen}{RGB}{34,139,34} 
\definecolor{softred}{RGB}{178,34,34}   

\usepackage{soul}
\newcommand{\hlc}[2][yellow]{{%
    \colorlet{foo}{#1}%
    \sethlcolor{foo}\hl{#2}}%
}

\usepackage{newfloat}
\usepackage{listings}
\DeclareCaptionStyle{ruled}{labelfont=normalfont,labelsep=colon,strut=off} 
\floatstyle{ruled}
\newfloat{listing}{tb}{lst}{}
\floatname{listing}{Listing}
\title{CRASH: Challenging Reinforcement-Learning Based Adversarial Scenarios For Safety Hardening}
\author{
    Amar Kulkarni,
    Shangtong Zhang,
    Madhur Behl
}
\affiliations{
    University of Virginia\\
    \{ark8su, shangtong, madhur.behl\}@virginia.edu
}

\begin{document}

\maketitle

\begin{abstract}
\input{abstract}
\end{abstract}

\section{1. Introduction}\label{sec:Motivation}
\input{introduction_mb}

\section{2. Related Work}\label{sec:RelatedWork}

\input{related_work}

\section{3. Problem Setup and Formulation}\label{sec:ProblemFormulation}
\input{problem_setup}

\section{4. CRASH: Adversarial Falsification and Safety Hardening}\label{sec:Methodology}
\input{methodology}

\section{5. Results}\label{sec:Results}
\input{results}

\section{6. Conclusions and Discussion}\label{sec:Conclusion}
\input{conclusion}

\clearpage
\bibliography{aaai25}

\end{document}

%% file: abstract.tex
Ensuring the safety of autonomous vehicles (AVs) requires identifying rare but critical failure cases that on-road testing alone cannot discover.  
High-fidelity simulations provide a scalable alternative, but automatically generating realistic and diverse traffic scenarios that can effectively stress test AV motion planners remains a key challenge. 
This paper introduces CRASH - Challenging Reinforcement-learning based Adversarial scenarios for Safety Hardening - an adversarial deep reinforcement learning framework to address this issue. 
First CRASH can control adversarial Non Player Character (NPC) agents in an AV simulator to automatically induce collisions with the Ego vehicle, falsifying its motion planner. 
We also propose a novel approach, that we term safety hardening, which iteratively refines the motion planner by simulating improvement scenarios against adversarial agents, leveraging the failure cases to strengthen the AV stack. 
CRASH is evaluated on a simplified two-lane highway scenario, demonstrating its ability to falsify both rule-based and learning-based planners with collision rates exceeding 90\%. 
Additionally, safety hardening reduces the Ego vehicle’s collision rate by 26\%. 
While preliminary, these results highlight RL-based safety hardening as a promising approach for scenario-driven simulation testing for autonomous vehicles.

%% file: introduction_mb.tex
Autonomous vehicles (AVs) are increasingly being deployed in cities, 
showcasing the potential of AV technology \cite{Kusano2023ComparisonOW}. 
Yet, these systems are far from perfect. 
Failures continue to occur, ranging from AVs getting stuck in traffic leading to frustrated commuters; all the way to catastrophic events with life-threatening consequences. 
A stark example is the recent Cruise pedestrian collision in San Francisco, where a pedestrian was struck by a human driven vehicle and flung in to the path of the AV which further dragged the pedestrian, leading to public outrage and the suspension of Cruise's operating license \cite{Exponent2024CruiseReport}. 
These failures are not isolated incidents-they expose a deeper issue. 
AVs struggle with rare and complex situations, known as ``long-tail'' events, which lie beyond the scope of everyday driving scenarios. 
Compounding the problem is a pervasive lack of transparency. 
When accidents occur, little to no information is shared about their root causes and whether meaningful steps are being taken to prevent similar failures in the future. 
Reports from the California DMV and NHTSA~\cite{Team2021, NHTSA2024ADSReport12345} provide limited insights, often describing only surface-level details. 
\begin{figure}
    \centering
     \includegraphics[width=\linewidth]{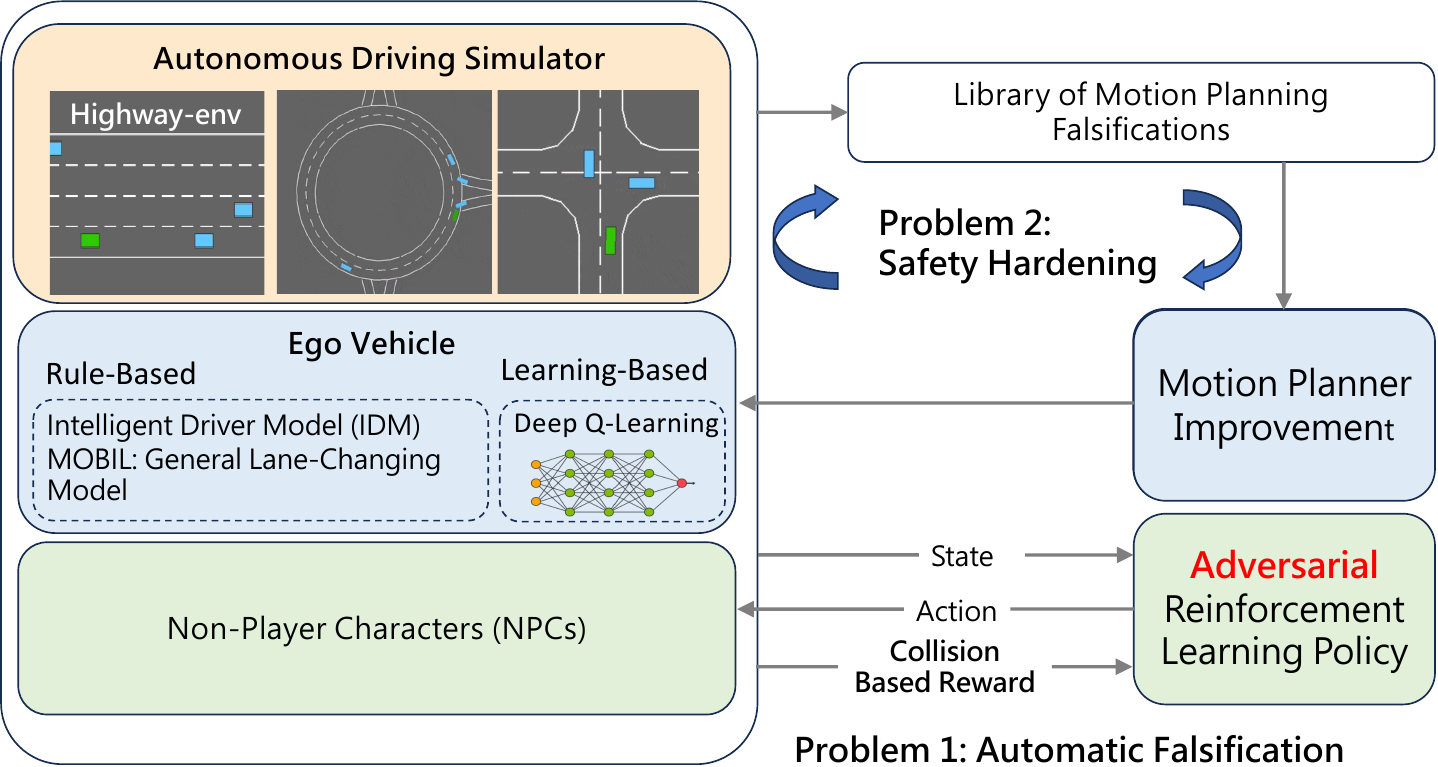}
     \caption{\textbf{CRASH} is a framework that iteratively tests and improves an AV motion planner. First, Automatic Falsification uses adversarial NPCs, guided by a reinforcement learning policy designed to induce collisions with the Ego vehicle. Then, Safety Hardening retrains the Ego motion planner to enhance its robustness against these adversarial scenarios.}
     \label{fig:hero}
 \end{figure}
The current approach to AV safety development relies heavily on manual post-incident analysis, where developers retrospectively analyze failure scenarios and craft expert-designed rules to mitigate similar incidents. 
This process is not only reactive but also highly inefficient, requiring vast amounts of driving time to encounter rare safety-critical situations that expose system vulnerabilities~\cite{KALRA2016182}. 
Moreover, the reliance on on-road testing raises ethical concerns, as cities and public roads are effectively used as testing grounds, putting lives at risk.

Consequentially, simulation has become a critical tool for addressing challenges in AV testing, providing a controlled environment for exploring rare safety-critical scenarios without risking lives. 
However, generating meaningful scenarios is complex and requires addressing three major challenges: \textit{saliency}, ensuring scenarios are realistic and relevant by avoiding unrealistic dynamics or overly simple interactions that fail to uncover weaknesses; \textit{curse of dimensionality}, managing the vast configuration space of variables like vehicle behaviors, road layouts, and environmental factors, which grow exponentially and complicate prioritization; and \textit{curse of rarity}, efficiently identifying critical long-tail scenarios that are infrequent but essential for comprehensive testing~\cite{nhtsaFatalityAnalysis}. 
Simulation must overcome these challenges of scenario generation to effectively target and evaluate meaningful edge cases.

A targeted approach is needed to simulate scenarios where AVs are likely to fail. 
While researchers have explored methods such as perturbing logged trajectories, recreating real-world accidents in simulation, and importance sampling to search for edge cases, these techniques often focus solely on identifying failure scenarios. 
Critically, they do not incorporate these scenarios into the process of improving the AV’s safety.

To address this gap, we propose CRASH, a novel framework that combines Automatic Falsification and Safety Hardening to systematically test and improve AV motion planners (shown in Figure~\ref{fig:hero}).
CRASH not only identifies failure-inducing scenarios through adversarial reinforcement learning but also integrates them into an iterative process to enhance the Ego vehicle’s robustness. 
This work focuses exclusively on motion planning, excluding considerations of perception in scenario generation.
The contributions of this paper are as follows:
\begin{enumerate}
    \item We develop a DQN-based method to train NPC agents capable of automatically falsifying both rule-based and learning-based Ego motion planners.
    \item We introduce the novel concept of Safety Hardening, formulated as a bi-level optimization, and we investigate three methods to improve the safety of Ego motion planners, including local hardening and model pool-based methods with uniform and prioritized sampling.
\end{enumerate}
Above all, this work attempts to lay the foundation for a unified methodology that addresses inefficiency and limitations in failure scenario generation and establishes a pathway for systematically improving AV safety through simulation.

%% file: related_work.tex
AV safety is a multifaceted problem, but surprisingly, there are no agreed-upon safety metrics.
Standards like ISO 26262 \cite{iso26262} focus on functional safety, addressing risks from software errors, and hardware faults, but they do not account for the planning or decision-making capabilities of AVs. 
Similarly, UL 4600 \cite{laboratories_ul_2020} provides a framework for creating safety cases but struggles with the challenge of enumerating all possible traffic scenarios. 
Formal methods for defining AV safety face challenges from the uncertainty and complexity of traffic environments \cite{MEHDIPOUR2023110692}, making scenario parametrization tedious and verification impractical for dynamic, unbounded scenarios. Rather than providing formal guarantees, we focus on evaluating safety through collision rates, a widely accepted and practical metric in autonomous driving \cite{criticality2021}.

Many simulation-based scenario generation studies rely on naturalistic driving data (NDD) to construct datasets of realistic scenarios, which can then be searched for dangerous events \cite{sun2021corner, neelofar_identifying_2022}. 
NDD also aids in building learning-based behavior models for vehicles and pedestrians \cite{yan_learning_2023}. 
The Naturalistic Adversarial Driving Environment (NADE), combines decision trees generated from NDD with deep reinforcement learning (DRL) to generate rare scenarios \cite{feng_dense_2023, feng_intelligent_2021}. 
While NDD ensures realism, it limits the scope of generated scenarios to maneuvers observed in the dataset, potentially missing the full dynamic capabilities of AVs. In contrast, our method directly generates collision trajectories, avoiding reliance on pre-existing data.

Actively probing systems for vulnerabilities is well-researched in autonomous driving. 
For instance, Robust Adversarial Reinforcement Learning (RARL) applies adversarial disturbances to improve controller robustness~\cite{pinto2017robust}, with variants like Risk Averse RARL~\cite{pan2019risk} focusing on vehicle control, whereas we focus on motion planning. 

\vspace{-4pt}
\begin{table}[h]
\centering
\caption{Comparison of CRASH with closest related work}
\vspace{-8pt}
\label{tab:comparison}
\resizebox{\columnwidth}{!}{%
\setlength{\tabcolsep}{4pt} 
\renewcommand{\arraystretch}{0.8}
\begin{tabular}{l|c|c|c|c}
\toprule
Feature              & NADE & RARL & STRIVE & CRASH \\ \midrule
Adversarial Generation        & \checkmark             & \checkmark             & \checkmark               & \checkmark              \\
Iterative Improvement         & $\times$              & $\times$              & $\times$                & \checkmark              \\
Focus on Motion Planning      & \checkmark             & $\times$              & \checkmark               & \checkmark              \\
Naturalistic Data Dependence  & High          & None          & High            & None           \\ \bottomrule
\end{tabular}%
}
\vspace{-10pt}
\end{table}
Methods in adversarial trajectory generation control NPCs using techniques such as Particle Swarm Optimization~\cite{adversarialtrajectory} or Monte Carlo Tree Search~\cite{klischat_falsifying_2023}.
STRIVE optimizes adversarial trajectories in the latent space of VAE model \cite{rempe2022strive}.
While some approaches employ DRL, they typically stop after a single iteration of AV improvement \cite{Sharif_2022, karunakaran_critical_2022}. 
While these focus solely on falsification, CRASH iteratively discovers failure cases and uses them to enhance AV robustness.
To the best of our knowledge, this is the first framework to integrate both adversarial scenario generation and safety improvement into a unified methodology (Table~\ref{tab:comparison}).

%% file: problem_setup.tex
\noindent \textbf{Problem Setup:}
We model the traffic scenario as a Markov Decision Process (MDP), a framework well-suited for decision-making in dynamic, stochastic environments such as traffic. 
The MDP is described by the tuple $(S, S_0, A, P, R)$, where $S$ is the state space, $S_0$ is the set of intial states, $A$ the action space, $P$ the transition model, and $R$ the reward function. 
The probabilistic transition relation $P: S \times A \times S \rightarrow \mathbb{R}$ defines the probability of transitioning from one state to another given an action, governed by the simulator dynamics.
The state of each vehicle $s_t^i \in S_t$ at time $t$ is represented by its position $(x_t, y_t)$, heading $\psi_t$, and velocity $v_t$ along the heading direction. 
\begin{figure*}
    \centering
     \includegraphics[width=\linewidth]{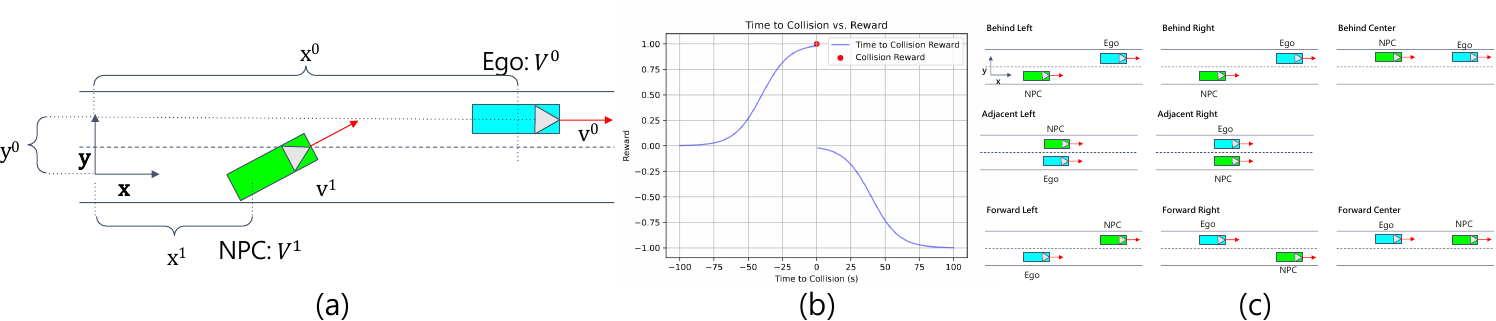}
     \caption{(a) Coordinate system: the blue vehicle ($V_0$) is the Ego, and the green vehicle ($V_1$) is an NPC on a 2-lane highway. (b) Reward function: positive when the NPC approaches the Ego, negative when moving away. (c) Initial configurations: the NPC (green) spawns at 8 relative positions around the Ego (blue).}
     \label{fig:coordinatesystem}
 \end{figure*}
The scenario consists of $N+1$ vehicles, where the Ego vehicle $V^0$ is controlled by the motion planner, and the remaining $N$ vehicles are Non-Player Characters (NPCs), denoted by $V^i$. 
Figure \ref{fig:coordinatesystem}(a) illustrates a reference coordinate system for a 2-lane, 2-agent scenario, showing the positions and headings of the Ego and an NPC vehicle.
Figure \ref{fig:coordinatesystem}(c) shows the example of the set of intial states $S_0$ for the same configuration comprising of 8 possible initial configurations between the Ego and NPC vehicles, based on their relative positions.
The motion planning for the Ego agent $V^0$, can be dictated by either a rule-based motion planner, or a learning-based motion planner (e.g. DQN)
For the learning-based planner, the reward function encourages collision avoidance and forward motion~\cite{highway-env}.
This setup provides the foundation for formulating the two key problems addressed in this paper.

\subsubsection{Problem 1: Automatic Falsification}
The goal of automatic falsification is to discover failure-inducing scenarios by optimizing the behavior of NPC agents to maximize collisions with the Ego vehicle. 
This involves finding an optimal policy $\pi^{*}_{\text{NPC}}(S_t, A_t)$ that maps states $S_{t}$ to actions $A_t$ to maximize the cumulative collision count: 
\begin{equation}
\label{eq:npc_policy}
    \pi^{*}_{\text{NPC}}(S_{t}, A_{t}) = \arg\max_{A_{t}}(\sum^{E}_{j=0}\phi)
\end{equation}
Here, $\phi$ is a binary indicator variable set to $1$ whenever a collision is detected by the simulator and $0$ otherwise, and $E$ denotes the total number of episodes. 
The simulator evaluates interactions between the NPC agents and the Ego vehicle, updating $\phi$ for each collision.

\subsubsection{Problem 2: Safety Hardening}
The outcome of a successful automatic falsification is an adversarial NPC agent that can falsify the Ego vehicle and result in several collisions. 
Given an adversarial NPC agent, the objective is to try and find an optimal policy $\pi^{*}_{\text{Ego}}(S_t, A_t)$ for the Ego, that minimizes the cumulative number of crashes across $E$ episodes. 
\begin{equation}
\label{eq:ego_policy}
    \pi^{*}_{\text{Ego}}(S_{t}, A_{t}) = \arg\min_{A_{t}}(\sum^{E}_{e=0}\phi)
\end{equation}
Notably, the optimization objectives for the Ego and NPC policies, $\pi^{}_{\text{Ego}}$ and $\pi^{}_{\text{NPC}}$, are dual problems: while the NPC aims to maximize collisions (Eq. \eqref{eq:npc_policy}), the Ego aims to minimize them (Eq. \eqref{eq:ego_policy}).
Safety hardening combines these dual objectives into a unified framework as a bi-level optimization problem, where the Ego and NPC policies are optimized sequentially. This is represented as:

\begin{equation}
    \label{eq:optimal_policy}
    \min_{\sum^{E}_{e=0}\phi} \max_{\sum^{E}_{e=0}\phi} f(\pi^{*}_{\text{Ego}}(S_{t}, A_{t}), \pi^{*}_{\text{NPC}}(S_{t}, A_{t}))
\end{equation}

Here, $f$ is a function that evaluates both policies, with the NPC attempting to exploit the Ego’s weaknesses while the Ego adapts to improve its robustness. 
This setup is similar to the adversarial training structure of Generative Adversarial Networks (GANs) \cite{Goodfellow2014GenerativeAN}, where one agent (NPC) generates adversarial scenarios and the other (Ego) learns to counteract them.

We make the following assumptions in our problems:
The state transition model $P_t$ is treated as a black-box, governed by the simulation’s underlying dynamics. The state $S_{t}^{i}$ of all vehicles is assumed to be fully known, without sensor noise or uncertainty. The set of all possible initial states is finite and enumerable. Finally, the Ego vehicle can be controlled using either a rule-based or learning-based motion planner, with the latter assumed to be re-optimizable and re-trainable.

%% file: methodology.tex
CRASH operates in two key stages: Automatic Falsification, where we train Non-Player Character (NPC) agents to discover failure-inducing scenarios for the Ego vehicle, and Safety Hardening, where the Ego motion planner is iteratively retrained to counter these adversarial scenarios, improving its robustness.
We describe each of these next.

\subsection{4.1 DQN Based Automatic Falsification}\label{subsec:falsification}

In simulation-based autonomous vehicle research, reinforcement learning is typically employed to improve the behavior of the Ego vehicle. 
However, in the context of falsification, we shift focus: rather than controlling the Ego, we are given an Ego motion planner and aim to train an NPC agent to learn a policy that maximizes collisions with the Ego. This adversarial setup forms the foundation of the automatic falsification in CRASH, where we leverage Deep Q-Networks (DQN)~\cite{mnih_human-level_2015} to train the NPC.\@

\subsubsection{Adversarial Reward Design:}
To incentivize the NPC to collide with the Ego, we design an adversarial reward function tailored to this objective. 
The total reward $R_t$ combines multiple components to guide the NPC’s behavior:
\begin{equation} 
\label{eq:reward}
    r^{i}_{t}(s^{i}_{t}, a^{i}_{t})= w_{1}r_{\text{c}} + w_{2}r_{\text{x}} + w_{3}r_{\text{y}}
\end{equation}
Here, $r_c$, $r_x$, and $r_y$ serve distinct roles:
$r_c$ is the boolean collision reward set to $1$ when a collision $\phi$ occurs, providing the primary signal for adversarial behavior. 
However, since collisions terminate the episode, $r_c$ is a sparse reward, presenting challenges for reinforcement learning~\cite{rengarajan2022reinforcement}.
To address the sparsity of $r_c$, we incorporate continuous rewards based on time-to-collision (TTC) along the longitudinal ($x$) and lateral ($y$) axes. These components, $r_x$ and $r_y$, reward the NPC for decreasing the distance to the Ego while penalizing motion away from it. 
The TTC, $\lambda$, is computed as: $\lambda_{x}=\frac{\Delta x}{\Delta v_x}$ and $\lambda_{y}=\frac{\Delta y}{\Delta v_y}$.
Unlike traditional TTC methods that focus only on positive values (collision proximity), we use the sign of $\lambda$ to encode the NPC’s approach direction. This enhances the NPC’s spatial awareness relative to the Ego.

To ensure smooth gradients for learning, $r_x$ and $r_y$ are computed using sigmoid functions applied to  $\lambda^i_{t,x}$  and  $\lambda^i_{t,y}$.  
 \begin{figure}
    \centering
     \includegraphics[width=0.5\textwidth]{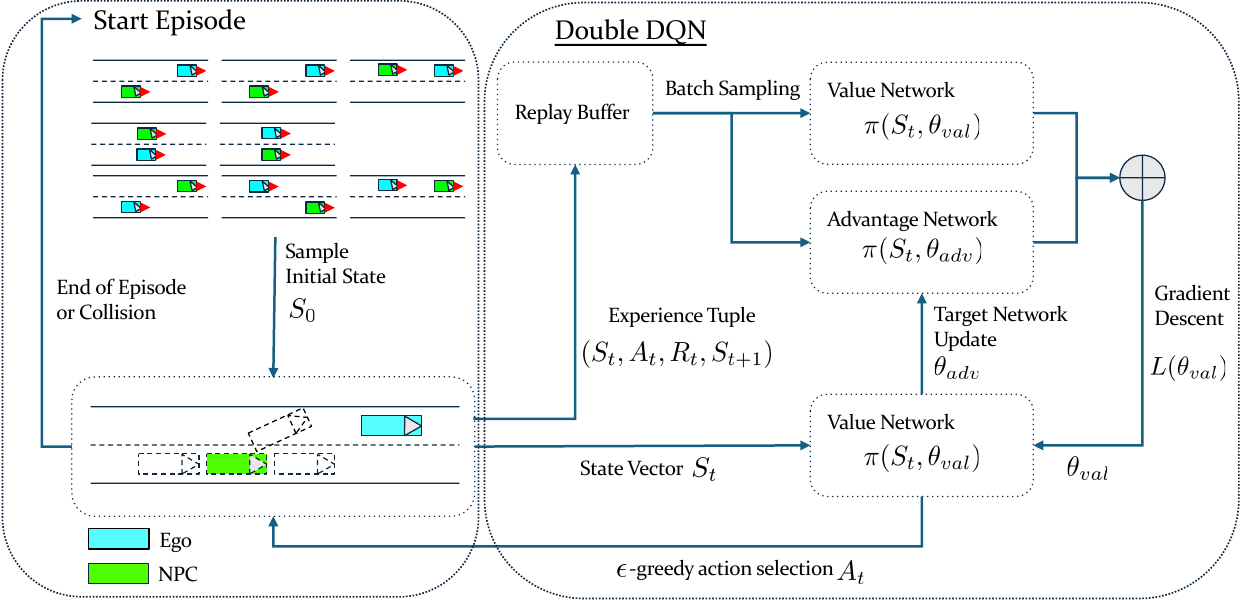}
     \caption{\textbf{Automatic Falsification in CRASH}: Training architecture for the adversarial NPC agent using Double DQN:\@ simulator states ($S_t$) are input to the value network, which selects actions via the $\epsilon$-greedy method}\label{fig:dqn_overview}
 \end{figure}
\begin{equation}
\label{eq:rx_sigmoid}
    r_{x} = \dfrac{-\text{sign}(\lambda^{i}_{t,x})}{1 + e^{a - b\lambda^{i}_{t,x}}} \quad
    r_{y} = \dfrac{-\text{sign}(\lambda^{i}_{t,y})}{1 + e^{a - b\lambda^{i}_{t,y}}}
\end{equation}

Here, $a$ and $b$ are scaling parameters tuned to balance the reward’s sensitivity to distance and velocity. 
The weights $w_1$, $w_2$, and $w_3$ govern the relative importance of each reward component and are tuned through reward shaping. Figure~\ref{fig:coordinatesystem} (b) illustrates the continuous TTC-based rewards.

\subsubsection{DQN-Based NPC Training:}
With the adversarial reward function, $R_t$, defined, we train the NPC to approximate the optimal action-value function $Q^{*}(S_t, A_t)$. 
For stability and efficiency, we use Double Q-Learning~\cite{vanhasselt2015deep} with prioritized experience replay~\cite{schaul_prioritized_2016}.
The NPC training architecture, shown in Figure~\ref{fig:dqn_overview}, consists of two networks: A value network parameterized by $\theta_{\text{val}}$ to estimate $Q(S_t, A_t; \theta_{\text{val}})$; and a target network parameterized by $\theta_{\text{adv}}$ for stable updates.
The value network is trained by minimizing the Bellman loss: 

\begin{equation}
\begin{aligned} \label{eq:squaredloss}
L(\theta_{val}) = (R_{t} &+ \gamma \max_{a} Q^{*}(S_{t+1}, A_{t+1}; \theta_{\text{val}}) \\ &- Q(S_{t}, A_{t}; \theta_{\text{adv}}))^{2}
\end{aligned}
\end{equation}
To address overestimation bias in Q-learning, the target network parameters, $\theta_{adv}$, are updated using an exponential moving average:
$\theta_{\text{adv}} \leftarrow \tau \theta_{\text{val}} + (1 - \tau) \theta_{\text{adv}}
$
where $\tau$ is the smoothing factor.

The training process begins by uniformly sampling initial states $S_0$ to prevent overfitting to a narrow set of scenarios. 
At each time step $t$, the simulator provides the current state $S_t$, which is fed into the value network to compute Q-values for all possible actions. An action $A_t$ is selected using an $\epsilon$-greedy strategy to balance exploration and exploitation, executed in the simulator, and results in a reward $R_t$ and next state $S_{t+1}$. 
The experience tuple $(S_t, A_t, R_t, S_{t+1})$ is stored in a prioritized replay buffer.
Once the buffer reaches a certain capacity, experiences are sampled to update the value network using the loss function in Equation~\eqref{eq:squaredloss}. 
During evaluation, actions are selected greedily based on the highest Q-value, without exploration. Trajectories resulting in collisions are stored as falsifications, collected during both training and evaluation.
This approach allows the NPC to systematically discover failure modes in the Ego motion planner.

\subsection{4.2 Safety Hardening}
\label{sec:SafetyHardening}
Building on the automatic falsification process, Safety Hardening aims to enhance the Ego agent’s safety by iteratively training it against the adversarial NPC policies. 
After training the optimal NPC $\pi^{*}_{\text{NPC}}(S_{t},A_{t})$ to cause collisions against the current Ego policy, $\pi_{\text{Ego}}(S_{t},A_{t})$, we train the Ego policy $\pi^{*}_{\text{Ego}}(S_{t},A_{t})$ to mitigate these failures.
We sequentially train $\pi_{\text{NPC}}$ and $\pi_{\text{Ego}}$, one at a time, over a fixed number of cycles $C$. 
Each cycle comprises of a Falsification simulation, during which the NPC is trained to improve $\pi_{\text{NPC}}$ by updating the NPC policy parameters $\theta_{\text{npc},c}$. 
This is followed by a Hardening simulation where the weights of the NPC DQN are frozen and the Ego policy parameters $\theta_{\text{ego},c}$ are updated to improve $\pi_{\text{Ego}}$ to learn to avoid collisions against an adversarial NPC.
We denote $V_{c} = \pi_{\text{NPC}}(S_t, A_t; \theta_{\text{npc},c})$ as the NPCs trained in the falsification simulation in cycle $c$ and $E_{c} = \pi_{\text{Ego}}(S_t, A_t; \theta_{\text{ego},c})$ as the Ego trained during the hardening simulation in cycle c.

\subsubsection{Local Safety Hardening} 
\label{sec:local_falsification_hardening}

\begin{figure}
    \centering
     \includegraphics[width=\linewidth]{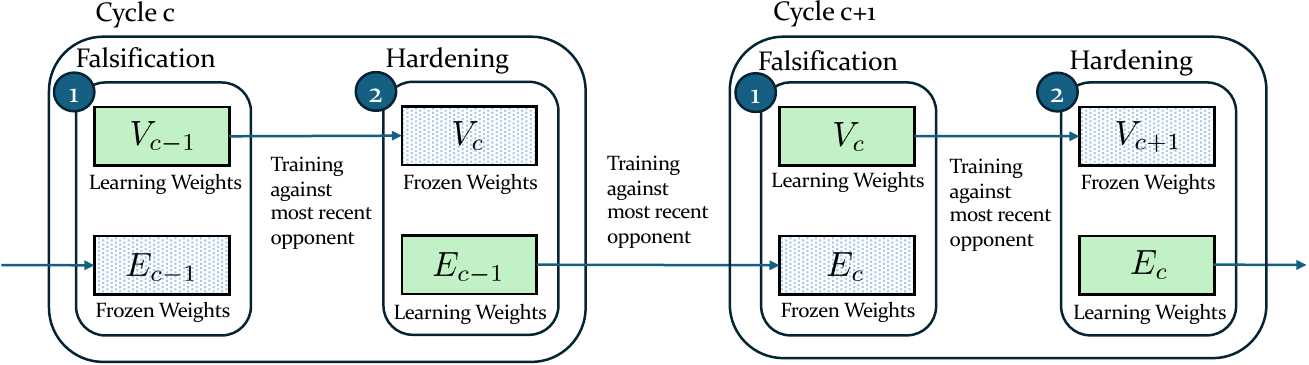}
     \caption{Local Safety Hardening over two cycles: In cycle $c$, the NPC $V_{c-1}$ (green) is trained against the fixed Ego $E_{c-1}$ (blue), producing $V_c$. Then, $V_c$ is fixed, and $E_{c-1}$ is trained to produce $E_c$. This process repeats in cycle $c+1$}
     \label{fig:localsh}
 \end{figure}

The simplest approach to Safety Hardening is Local Safety Hardening, where each NPC is trained exclusively against the most recent Ego agent from the previous cycle. 
As illustrated in Figure~\ref{fig:localsh}, during cycle $c$, the NPC agent $V_c$ is trained in a  Falsification simulation to maximize collisions with $E_{c-1}$, whose weights remain fixed. 
Following this, $E_c$ is trained during a Hardening simulation to minimize collisions with $V_c$, whose weights are now frozen. 
This alternation between falsification and hardening repeats for each cycle.
While straightforward, Local Safety Hardening can lead to overfitting, as agents are optimized only against their immediate counterparts, limiting their ability to generalize to a broader range of opponents.

\subsubsection{Uniform Model Pool-Based Safety Hardening:} \label{sec:unisampling}

To address the overfitting issue inherent in Local Safety Hardening, we introduce Model Pool-Based Safety Hardening, which maintains a pool of all previously trained agents. 
Specifically, we maintain two model-pools - one for the set of all previously trained Egos $E^{P} = \{E_{c}\}, \forall c \in C$ and another for the set of all NPCs $V^{P} = \{V_{c}\}, \forall c \in C$. 
After training, each agent’s weights are frozen, and it is added to the respective pool to preserve its behavior.
\begin{figure}[b]
    \centering
     \includegraphics[width=0.5\textwidth]{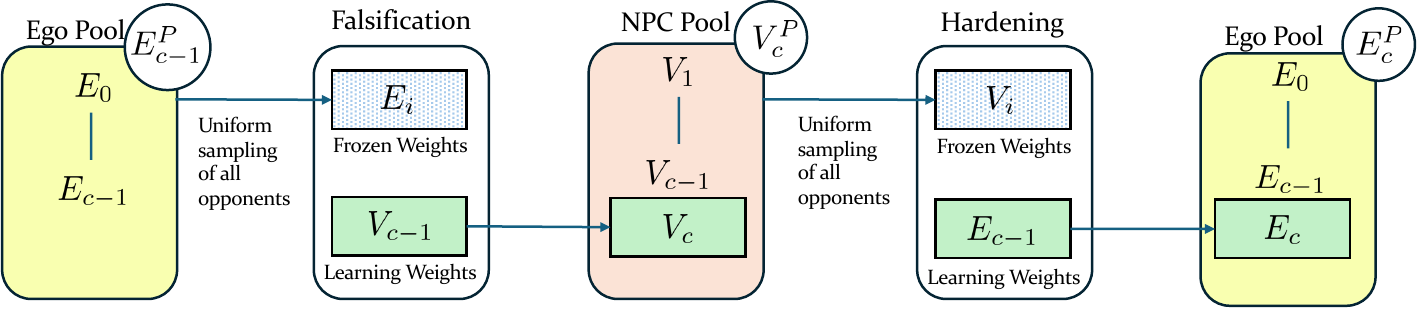}
     \caption{Uniform Sampling for model pool-based safety hardening: The NPC $V_{c-1}$ is trained against Ego agents sampled uniformly from the pool $E^P_{c-1}$, and the resulting NPC, $V_c$, is added to the NPC pool $V^P_{c}$. The newest Ego is then trained against NPCs sampled uniformly from $V^P_{c}$.}
     \label{fig:model_pool_uniform}
 \end{figure}
During Falsification and Hardening simulations, agents are sampled from the model pool to serve as opponents for the current episode. The key question is: what sampling criteria should be used? As the name suggests, uniform model pool-based sampling selects opposing agents uniformly from the corresponding pool.
Figure~\ref{fig:model_pool_uniform} depicts this process. 
For NPC training at cycle $c$, agents are sampled from the Ego pool $E^P_{c-1}$ using uniform probability, ensuring exposure to a diverse range of opponents and mitigating the risk of forgetting how to be effective against previous agents, whether that is for falsification in the case of NPCs, or for hardening in the case of the Ego vehicle.
Similarly, for Ego training at cycle $c$, agents are sampled uniformly from the NPC pool $V^P_{c}$. 
The uniform model pool-based sampling approach promotes generalization by exposing agents to varied strategies and reducing the risk of overfitting to a single opponent.

\subsubsection{Prioritized Model Pool-Based Safety Hardening:} \label{sec:priosampling}
While uniform sampling improves generalization, it effectively trains an agent to handle the average skill level of the pool rather than trying to iteratively building stronger agents. 
To address this, the prioritized model pool-based sampling can be employed, where agents are sampled based on their skill levels, determined by an Elo rating system \cite{Elo1967}. 
Each new NPC or Ego enters the pool and starts with an Elo rating of $R=1000$. 
The expected score $\mathbb{E}_{\text{Agent}}$ for an agent is computed using the rating difference with its opponent:
\begin{align}
    \mathbb{E}_{\text{Agent}} &= \frac{1}{1 + \exp\left( \dfrac{R_{\text{Opponent}} - R_{\text{Agent}}}{\zeta} \right)}
\label{eq:expected_score}
\end{align}
Here $\zeta$ is a scaling factor to control the sensitivity to rating differences.
The agent score is updated each episode, where NPCs win if a collision happens, and Egos win if not. 
An agents score is updated iteratively $R'_{\text{Agent}} = R_{\text{Agent}} + K(\phi - \mathbb{E}_{\text{Agent}})$, where $K$ is a gain.
The sampling probability for each agent in the model pool is determined based on its expected score. For agent  $A_i$ , the probability  $P(A_i)$  of being selected is given by:
\begin{align}
    P(A_i) = \frac{w(A_i)}{\sum_{j} w(A_j)}, \quad \text{where} \quad w(A_i) = \left[ \mathbb{E}_{A_i} \right]^\beta
    \label{eq:model_probability}
\end{align}
Here, $\beta$ controls prioritization strength, with higher values favoring stronger agents. 
This method ensures that NPCs and Egos are trained against the most challenging opponents, promoting improvement.
In our traffic scenario, outcomes are binary: win or a loss. 
To ensure Elo ratings accurately reflect performance, we define distinct win criteria for NPCs and Ego based on their respective objectives.

\subsubsection{Tournament Based Evaluation}
\label{sec:tourney}

\begin{figure}
    \centering
     \includegraphics[width=0.5\textwidth]{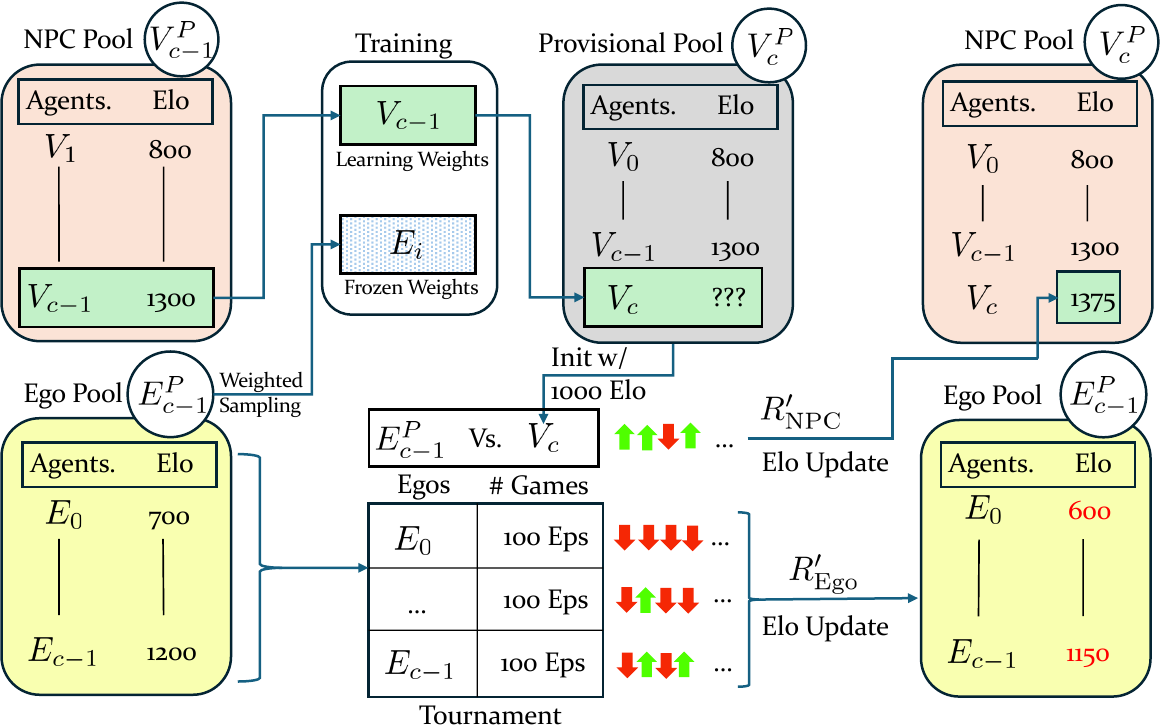}
     \caption{Example of $c$th safety hardening cycle with prioritized sampling and tournament evaluation: (1) NPC $V_{c-1}$ is trained against $E^P_{c-1}$, producing $V_c$. (2) $V_c$ is initialized with 1000 Elo and competes in a round-robin tournament with $E^P_{c-1}$ to update scores. (3) $V_c$ is added to the updated NPC pool $V^P_c$.}
     \label{fig:tournament_npc}
 \end{figure}

To mitigate biases introduced during training due to exploration (e.g., $\epsilon$-greedy actions), we implement a Tournament-Based Evaluation system. 
After training each new NPC or Ego agent, a round-robin tournament is conducted where the new agent competes against all agents in the opposing model pool (e.g., the new NPC $V_c$ competes against all Egos in $E^P$).
During these tournaments, agents act greedily without exploration, ensuring that their performance reflects true proficiency. 
Elo ratings are updated based on the tournament results, providing an accurate assessment of agent skill. 
Figure~\ref{fig:tournament_npc} illustrates this process for NPC evaluation. 
This procedure is similarly applied during Ego training, ensuring that the updated pools $V^P$ and $E^P$ reflect accurate agent rankings and behaviors.

%% file: results.tex

\subsection{5.1 Experiment Setup}
\label{sec:experiment_setup}

We use the \texttt{highway-env} simulator \cite{highway-env}, a gymnasium environment for testing RL algorithms in traffic scenarios. The setup consists of 1 Ego vehicle ($V^0$) and 1 NPC ($V^1$) on a two-lane highway, controlled by either a rule-based or learning-based planner. 
Each episode starts from one of 8 initial configurations (Figure~\ref{fig:coordinatesystem}(c)), chosen uniformly.
The rule-based planner for the Ego uses the Intelligent Driver Model (IDM) \cite{Treiber_2000} for longitudinal control and MOBIL \cite{mobil} for lane changes. 
The learning-based planner employs a DQN, while NPC agents use the reward function in Eq~\eqref{eq:reward} to prioritize causing collisions and minimizing time-to-collision. 
Both value and advantage networks are 3-layer perceptrons with 256 hidden units. 
The Adam optimizer \cite{Kingma2014AdamAM} is used with a learning rate of $5 \times 10^{-4}$. 
We primarily measure the average Crash Rate($CR$) over the last 100 episodes $CR=\sum_{100}^{i=0}\frac{\phi_{i}}{100}$, as well as the accumulated reward over each episode.

\subsection{5.2 Automatic Falsification Results}
\label{sec:FalsificationResults}

\begin{figure}
\centering
  \begin{subfigure}{\columnwidth}
    \centering
    \includegraphics[width=0.9\columnwidth]{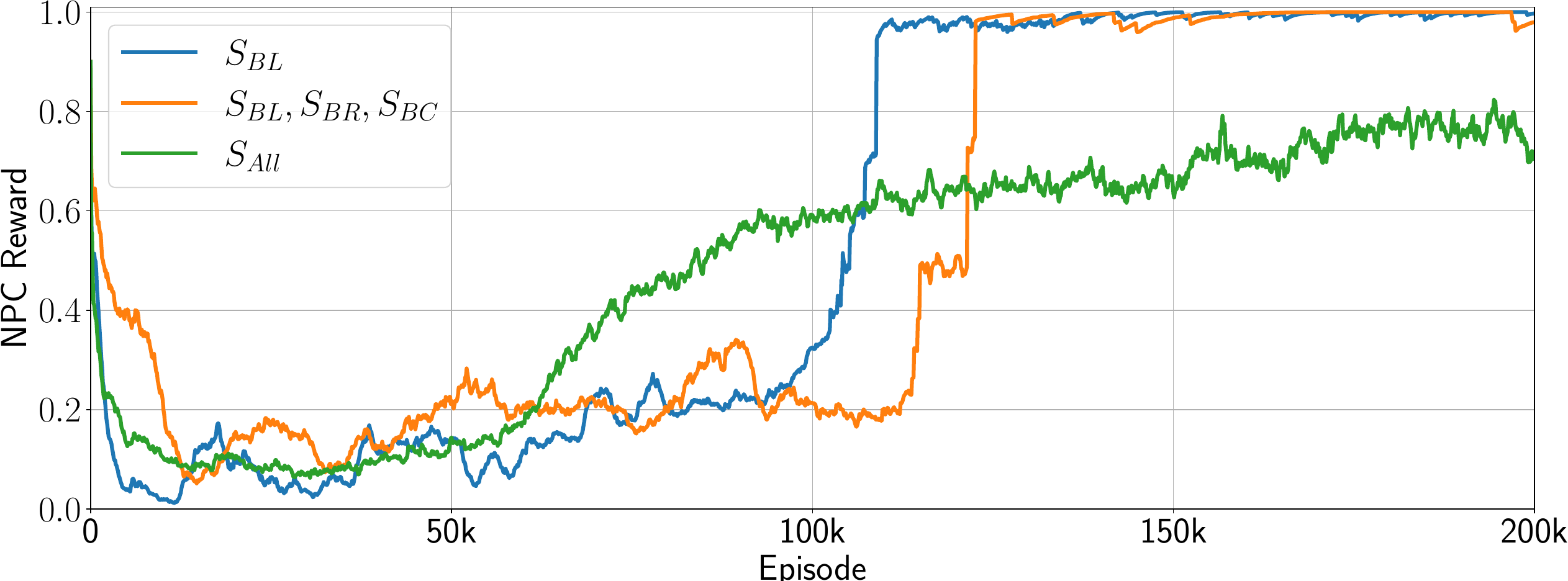}
    
  \end{subfigure}
  \vfill
  \begin{subfigure}{\columnwidth}
    \centering
    \includegraphics[width=0.9\columnwidth]{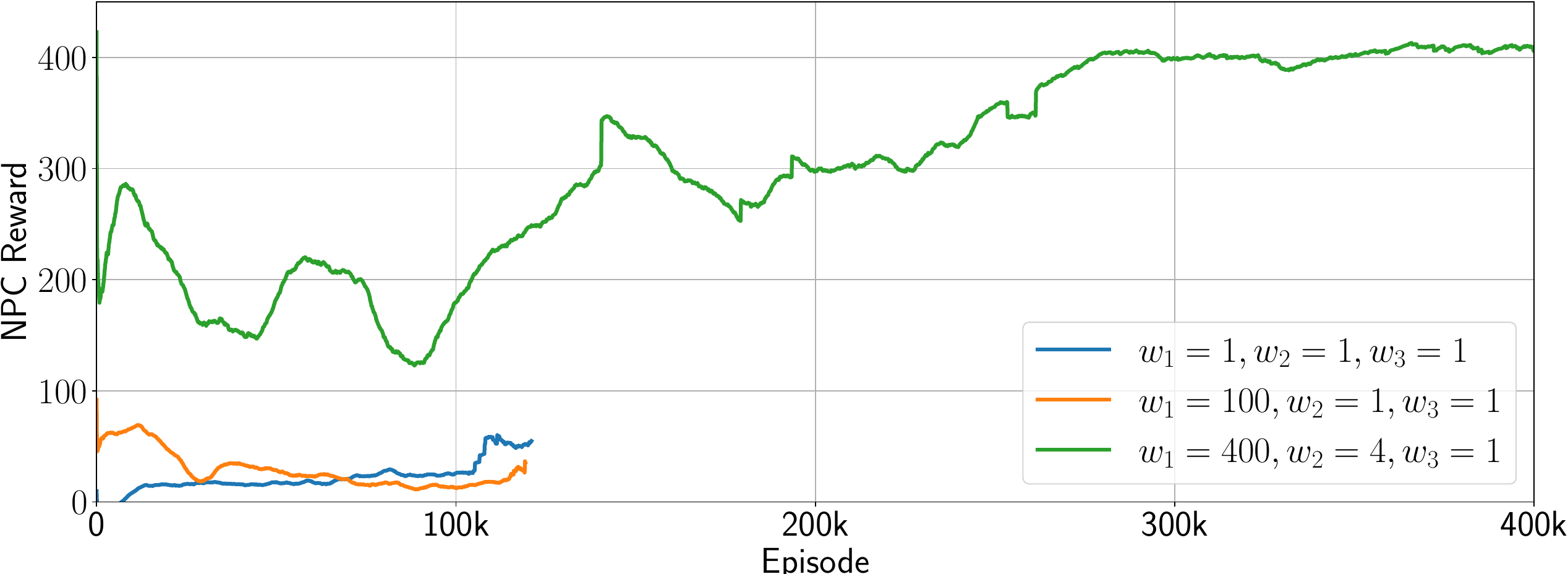}
  \end{subfigure}
  \caption{Accumulated rewards for NPC across sparse [Top] and continuous [Bottom] reward shaping experiments.}
  \label{fig:weights}
\end{figure}

To evaluate the effectiveness of reward shaping for the NPC adversarial agent, we examined its performance using sparse and continuous rewards to falsify the rule-based Ego motion planner. 
Figure~\ref{fig:weights} illustrates the accumulated rewards for the NPC across these two experiments: sparse rewards [Top] and continuous rewards [Bottom].
Initially, only the sparse reward is used, where $w_2$ and $w_3$ in Eq~\eqref{eq:reward} are set to zero. 
We test the sparse reward’s effectiveness under varying initial configurations: (1) the NPC spawns behind and to the left of the Ego ($S_0 = {S^{BL}}$), (2) the NPC spawns in any “behind” state ($S_0 = {S^{BL}, S^{BR}, S^{BC}}$), and (3) the NPC spawns uniformly across all possible states. 
Results in Figure~\ref{fig:weights}[Top] show that in simple scenarios, where the NPC spawns behind the Ego, it quickly maximizes the collision reward by speeding up and causing a rear collision. 
However, as more initial states are introduced, the learning rate drops significantly, indicating that sparse rewards alone are insufficient for the NPC to learn effective falsification strategies across diverse scenarios.
Figure~\ref{fig:weights}[Bottom] presents results when the continuous time-to-collision reward is integrated into the reward function, with varying weights $w_1$, $w_2$, and $w_3$ governing the emphasis on sparse rewards ($w_1$), longitudinal rewards ($w_2$), and lateral rewards ($w_3$). 
These trials reveal that the reward weights $w_1 = 400$, $w_2 = 4$, and $w_3 = 1$ enable the most effective learning for the NPC across all scenarios. 
The weights are thus retained for all subsequent experiments.

\begin{figure}
  \begin{subfigure}[b]{0.49\linewidth}
    \includegraphics[width=\linewidth]{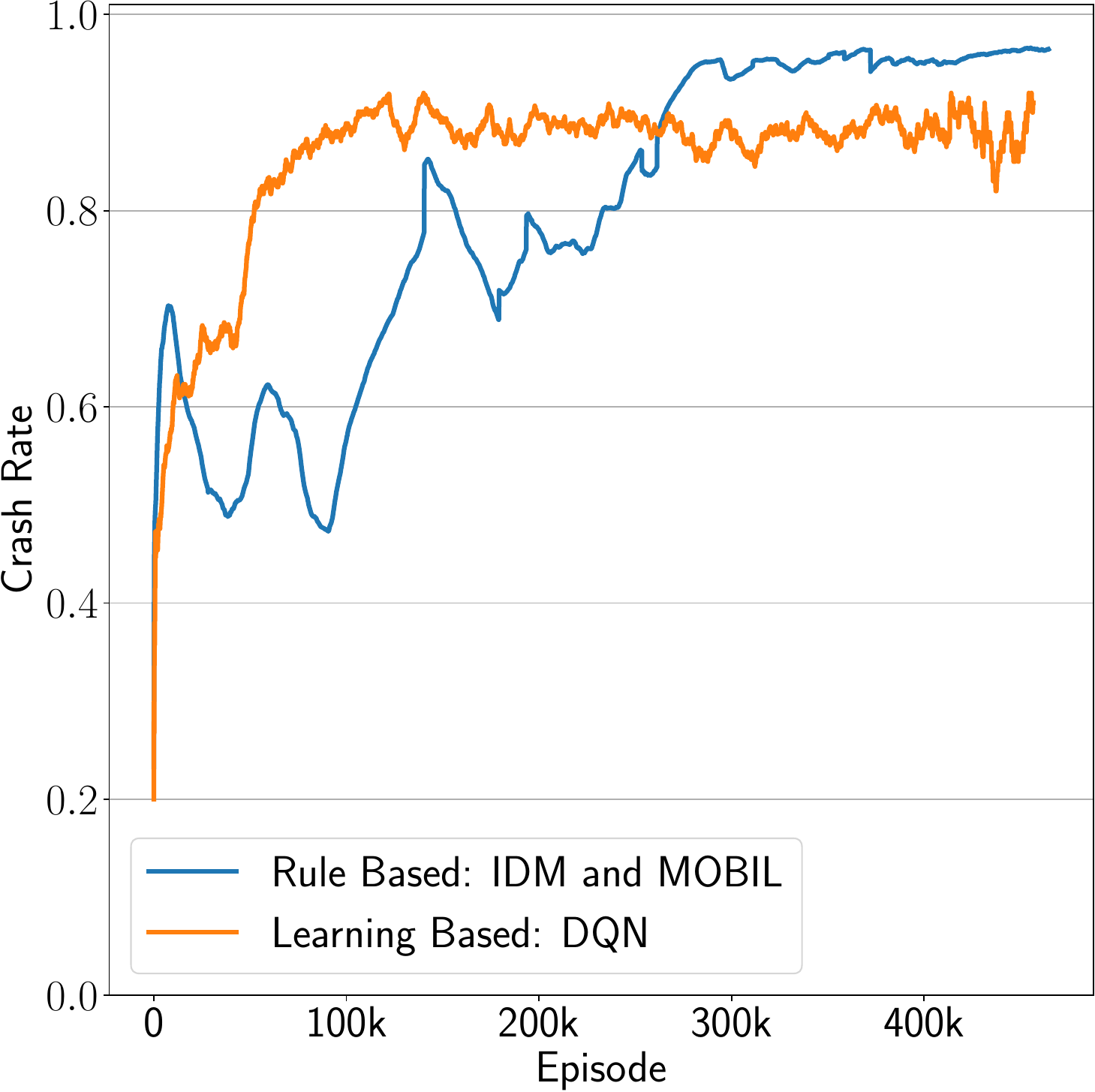}
    
  \end{subfigure}
  \hfill
  \begin{subfigure}[b]{0.49\linewidth}
    \includegraphics[width=\linewidth]{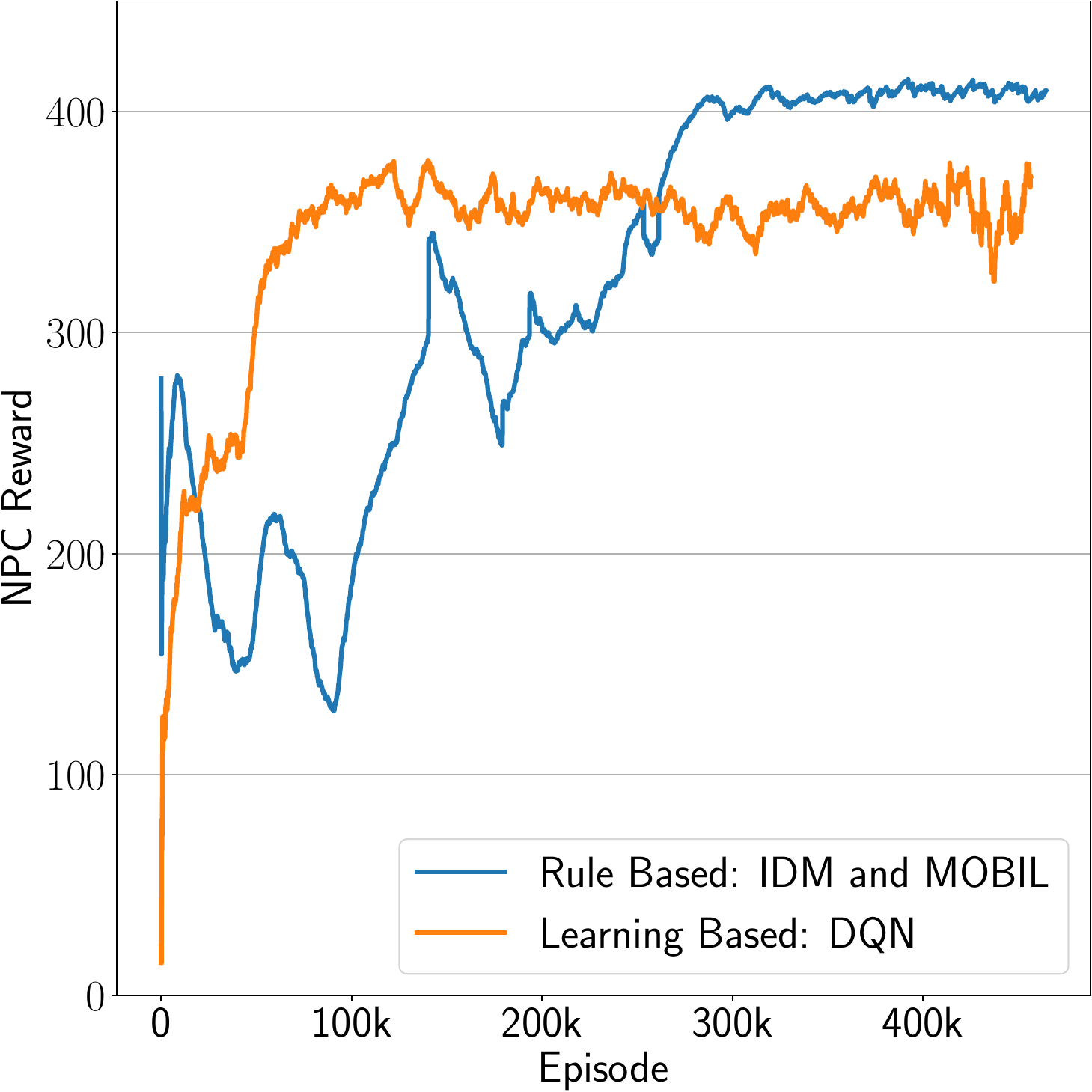}
  \end{subfigure}
  \caption{Average reward and crash rates for Automatic Falsification of rule-based and learning-based planners.}
  \label{fig:automatic_falsification_results}
\end{figure}

Using the tuned reward function, we trained the NPC to falsify both the rule-based and learning-based Ego motion planners. 
Ten trials, each with 10M transitions, were conducted, and the average crash rate and accumulated reward are shown in Figure~\ref{fig:automatic_falsification_results} [Left] and [Right], respectively. 
As the reward saturates, the NPC achieves an average crash rate of $97\%$ against the rule-based (IDM \& MOBIL) and $90\%$ against the learning-based (DQN) planner. 
These results highlight the effectiveness of the NPC reward function and the DQN implementation in Section~\ref{subsec:falsification} for automatic falsification of both planners in the two-lane, two-agent scenario.

\begin{table*}
  \centering
  \setlength{\tabcolsep}{4pt} 
\resizebox{\textwidth}{!}{
  \begin{tabular}{|l|ccc|ccc|ccc|ccc|ccc|ccc|}
  \hline
  & \multicolumn{3}{c|}{$V_1$} & \multicolumn{3}{c|}{$V_2$} & \multicolumn{3}{c|}{$V_3$} & \multicolumn{3}{c|}{$V_4$} & \multicolumn{3}{c|}{$V_5$} & \multicolumn{3}{c|}{Mean} \\ \hline
  & \scriptsize{Local} & \scriptsize{Uniform} & \scriptsize{Prioritized} & \scriptsize{Local} & \scriptsize{Uniform} & \scriptsize{Prioritized}  & \scriptsize{Local} & \scriptsize{Uniform} & \scriptsize{Prioritized}  & \scriptsize{Local} & \scriptsize{Uniform} & \scriptsize{Prioritized}  & \scriptsize{Local} & \scriptsize{Uniform} & \scriptsize{Prioritized} & \scriptsize{Local} & \scriptsize{Uniform} & \scriptsize{Prioritized}   \\ \hline
  $E_0$ & \cellcolor{LimeGreen}{$0.86 \uparrow$} & {0.80} & 0.83 & {0.79} & 0.74 & {0.67} & {0.36} & {0.40} & 0.37 & {0.28} & {0.67} & 0.49 & {0.69} & 0.57 & {0.44} & 0.60 & \textbf{{0.64}} & {0.56}\\ 
  
  $E_1$ & \cellcolor{SkyBlue}{$0.40 \downarrow$} & 0.32 & {0.24} & \cellcolor{LimeGreen}{$0.87 \uparrow$} & {0.38} & 0.76 & {0.61} & 0.47 & {0.38} & {0.64} & {0.49} & 0.58 & {0.38} & {0.48} & 0.43  & {0.58} & {0.43} & 0.48 \\ 
  
  $E_2$ & 0.32 & {0.24} & {0.38} & \cellcolor{SkyBlue}{$0.29 \downarrow$} & {0.31} & {0.27} & \cellcolor{LimeGreen}{$0.67 \uparrow$} & {0.44} & {0.44} & {0.17} & {0.47} & 0.36 & {0.21} & {0.44} & 0.32  & {0.33} & {0.38} & 0.35 \\ 
  
  $E_3$ & {0.66} & {0.38} & 0.56 & {0.75} & {0.32} & {0.75} & \cellcolor{SkyBlue}{$0.32 \downarrow$} & {0.38} & {0.28} & \cellcolor{LimeGreen}{$0.70 \uparrow$} & {0.58} & 0.63 & 0.50 & {0.57} & {0.36}  & {0.59} & {0.45} & 0.52 \\ 
  
  $E_4$ & {0.83} & {0.33} & 0.55 & {0.80} & {0.36} & 0.55 & {0.74} & {0.37} & 0.52 & \cellcolor{SkyBlue}{$0.17 \downarrow$} & {0.24} & {0.15} & \cellcolor{LimeGreen}{$0.77 \uparrow$} & {0.34} & 0.63 & {0.66} & {0.33} & 0.48\\ 
  
  $E_5$ & \cellcolor{Lavender}{0.62} & {0.31} & 0.52 & \cellcolor{Lavender}{0.58} & {0.43} & {0.92} & \cellcolor{Lavender}{0.61} & {0.28} & 0.56 & \cellcolor{Lavender}{0.58} & {0.50} & {0.63} & \cellcolor{SkyBlue}{$0.20 \downarrow$} & {0.39} & 0.26
   & 0.52 & \textbf{{0.38}} & {0.58}\\ \hline
  
  Mean &{0.62} & {0.40} & 0.51 & {0.68} & {0.42} & 0.65 & {0.55} & {0.39} & 0.43 & {0.42} & {0.49} & 0.47 & 0.46 & {0.47} & {0.41} & {0.49} & {0.43} & {0.49} \\ \hline
  \end{tabular}
  }
  \caption{Crash Rates of each Ego $E$ evaluated against each NPC $V$ for 100 episodes over 5 cycles of Falsification and Safety Hardening. Each method of Local Safety Hardening, Uniform model pool based sampling, and Prioritized model pool based sampling is shown.}
  \label{table:results_table}
\end{table*}

\subsection{5.3 Safety Hardening Results}
\label{sec:safety_results}

To evaluate Safety Hardening, we train the Ego and NPC policies iteratively over five cycles. 
Initially, the NPC policy $V_1$ is trained against the baseline Ego policy $E_0$. 
Subsequently, the Ego policy $E_1$ is retrained to counter $V_1$, and this process alternates through five cycles, with $V_{c}$ being trained against $E_{c-1}$ and $E_{c}$ against $V_{c}$.
We apply this training procedure to evaluate the three different Safety Hardening strategies: (1) Local Safety Hardening, 
(2) Uniform Model Pool-Based Safety Hardening, 
and (3) Prioritized Model Pool-Based Safety Hardening. 
Table~\ref{table:results_table} summarizes the results of these experiments. 
Each row corresponds to an Ego agent ($E_{c}$), showing its average crash rates evaluated over 100 episodes against the directly trained NPC counterpart ($V_{c}$) and all previously trained NPCs ($V_{1-4}$), for each method.
Similarly, each NPC ($V_{c}$) is evaluated against its immediate Ego counterpart ($E_{c-1}$) as well as all previously trained Egos ($E_{1-4}$). 
We first examine the trends in local falsification and hardening. 
In the first cycle, $V_1$ is trained against $E_0$, resulting in a crash rate of \hlc[LimeGreen]{$86\% \uparrow$}. 
Once $V_1$’s weights are frozen, $E_1$ is trained against $V_1$, reducing the crash rate to \hlc[SkyBlue]{$40\% \downarrow$}, demonstrating effective hardening. 
Subsequently, $V_2$ is trained against $E_1$ (with $E_1$’s weights frozen), increasing the crash rate to \hlc[LimeGreen]{$87\% \uparrow$}, indicating successful falsification. 
This is followed by $E_2$ reducing the crash rate to \hlc[SkyBlue]{$29\% \downarrow$} during hardening. These trends persist across cycles, with crash rates consistently increasing during falsification ($V_{c} \text{vs.} E_{c-1}$) and decreasing during hardening ($E_{c} \text{vs.} V_{c}$).
However, examining the final Ego agent, $E_5$, reveals limitations of local training. 
While $E_5$ achieves a low crash rate of \hlc[SkyBlue]{$20\% \downarrow$} against the most recent NPC, $V_5$, its crash rate against NPCs from earlier cycles ($V_{1-4}$) remains significantly higher ($>58\%$). This indicates that repeated training cycles did not build upon the skills learned in earlier iterations, and newly trained Egos tend to forget how to counter older NPCs. 
The mean crash rate further shows no consistent decrease across NPC versions, highlighting the inability of local falsification to generalize across iterations.

\begin{figure}
  \begin{subfigure}{0.49\linewidth}
    \includegraphics[width=\linewidth]{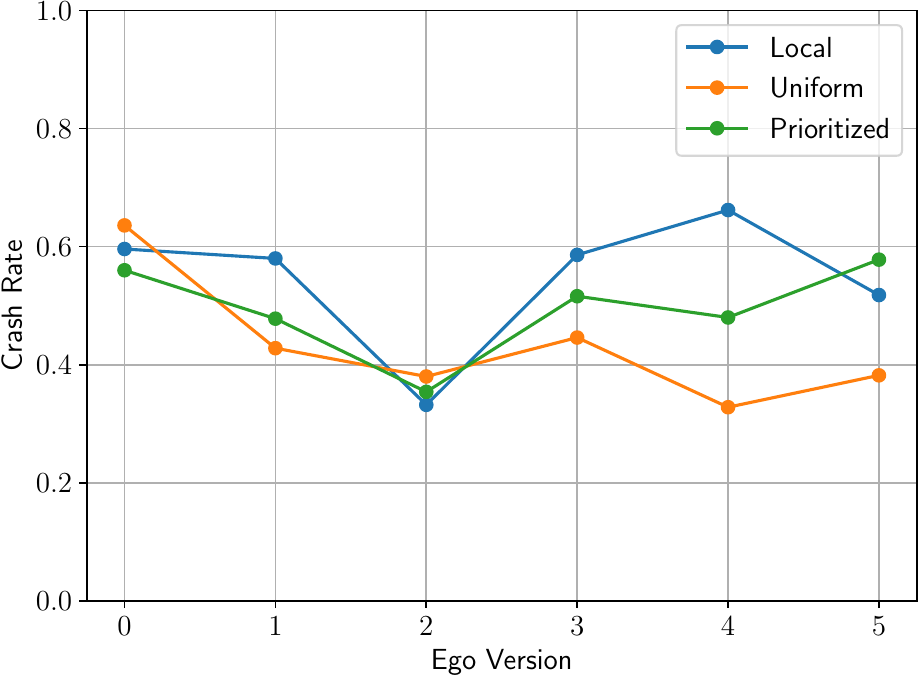}
    \caption{Mean Crash Rate across Ego Versions}
    \label{fig:avg_sh_sr_ego}
  \end{subfigure}
  \hfill
  \begin{subfigure}{0.49\linewidth}
    \includegraphics[width=\linewidth]{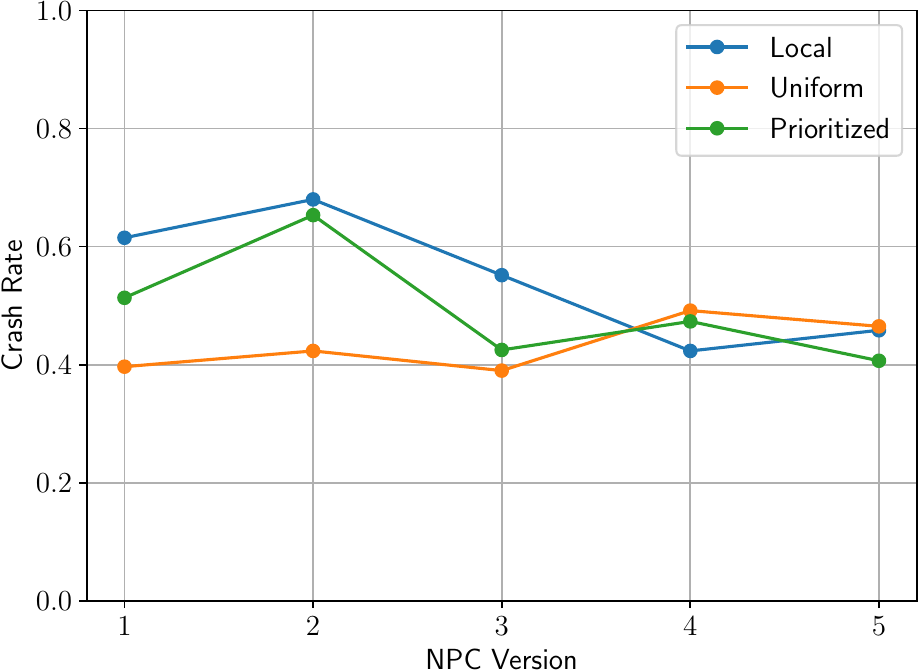}
    \caption{Mean Crash Rate across NPC Versions}
    \label{fig:avg_sh_sr_npc}
  \end{subfigure}
  \caption{The average crash rate across each variant of Safety Hardening. In the left plot, the average crash rate of all NPCs evaluated against each Ego version is shown. In the right plot, it is the average crash rate of all Egos evaluated against each NPC version.}
\label{fig:uniform_result}
\end{figure}

Figure~\ref{fig:uniform_result} illustrates the average crash rates for each Ego and NPC across cycles. 
Uniform model pool sampling exhibits the most consistent improvement, achieving a $26\%$ reduction (from $64\%$ to $38\%$) in the average crash rate for the Ego across five cycles, outperforming both local falsification and prioritized model pool sampling.
For NPC crash rates (Figure~\ref{fig:uniform_result}[Right]), no clear trend emerges to indicate which method is most effective. 
However, uniform model pool sampling shows the highest overall increase in NPC crash rates across the five cycles, suggesting it balances the effectiveness of both adversarial falsification and hardening better than the other methods.

%% file: conclusion.tex
In this work, we introduce CRASH, a novel framework that tackles the dual challenges of automatic falsification and safety hardening for AV motion planners. 
For automatic falsification, we demonstrate that a DQN-based adversarial NPC policy can effectively falsify both rule-based (crash rate $97\%$) and learning-based Ego motion planners (crash rate $90\%$) in a two-vehicle, two-lane highway scenario. 
Additionally, we propose three methods for iterative safety hardening, showing that the uniform model pool-based approach outperforms both local hardening and prioritized sampling. 
Specifically, it achieves a $26\%$ reduction in Ego crash rates over five hardening cycles, highlighting its effectiveness in mitigating overfitting and improving the generalization of trained agents.
In conclusion, this preliminary work shows the potential of the CRASH framework in developing adversarial agents for safety testing in autonomous driving and highlights the rich problem space in implementing safety hardening. 

\textbf{Limitations and Future Work:}
While our results demonstrate the efficacy of CRASH, we acknowledge the simplicity of the two-vehicle, two-lane highway scenario. 
For instance, for this scenario, the set of eight initial states can be fully enumerated a priori. 
Therefore, we anticipate when we scale this approach to include more NPCs and more complex road geometries, we will need to relax the assumption associated with the knowledge of all the initial states.  
Another problem of interest with multiple NPCs would be whether they can collude to cause more sophisticated falsifications for the Ego vehicle. 
This attribute is not examined yet by our approach and would likely necessitate exploring multi-agent deep reinforcement learning methods that could train all NPCs at once. 
Secondly, our current definition for falsification is simply that a collision occurs between the NPC and the Ego vehicle but not all collisions are equally informative; future investigations will explore collision-specific rewards to focus on generating failures where the AV is clearly at fault, offering more actionable insights into planner vulnerabilities. 
Despite these simplifications, our approach provides a valuable foundation for systematically testing and improving AV motion planners, addressing key challenges in safety-critical scenario generation and mitigation.

